\documentclass[11pt]{article}
\usepackage[]{ACL2023}

\usepackage{fancyhdr}
\fancypagestyle{plain}{%
  \fancyhf{}%
  \fancyhead[C]{Accepted to EACL 2026 Main Conference (Camera-Ready Version)}%
  \fancyfoot[C]{\thepage}%
}
\fancypagestyle{fancydefault}{%
  \fancyhf{}%
  \fancyhead[C]{Accepted to EACL 2026 Main Conference (Camera-Ready Version)}%
  \fancyfoot[C]{\thepage}%
}
\usepackage{tabularx}
\usepackage{booktabs}
\usepackage{xcolor}
\usepackage{xurl}
\usepackage{multirow}
\usepackage{caption}
\usepackage{pifont}
\usepackage{amsmath}
\usepackage{makecell}
\usepackage{graphicx}
\usepackage{subcaption}
\usepackage{times}
\usepackage{longtable}
\usepackage{tcolorbox} 
\usepackage{caption} 
\usepackage{float} 
\usepackage{microtype}
\usepackage{inconsolata}
\usepackage{arydshln} 
\usepackage{enumitem}
\usepackage{tikz}

\usepackage{placeins} 
\usepackage{changepage}
\usepackage{fancyvrb}

\usepackage[table]{xcolor}
\usepackage{colortbl}

\definecolor{metaleadblue}{HTML}{2f6490}

\DeclareCaptionType{Text Box}

\definecolor{customblue}{RGB}{33,167,234}
\definecolor{customgrey}{RGB}{240, 240, 240}
\definecolor{customorange}{RGB}{255, 127, 14}

\newcommand{\cmark}{\ding{51}}
\newcommand{\xmark}{\ding{55}}

\newfloat{textbox}{htb}{lop}[section] 
\floatname{textbox}{Textbox} 
\usepackage{setspace}

\definecolor{green}{named}{black} 

\title{\textsc{MetaLead}:
A Comprehensive Human-Curated Leaderboard Dataset for Transparent Reporting of Machine Learning Experiments}

\author{
Roelien C. Timmer \quad Necva Bölücü  \quad Stephen Wan\\ 
CSIRO Data61, Australia \\
\texttt{\{roelien.timmer, necva.bolucu, stephen.wan\}@data61.csiro.au} \\
}

\begin{document}

\maketitle

\thispagestyle{plain} 

\begin{abstract}
Leaderboards are crucial in the machine learning (ML) domain for benchmarking and tracking progress. However, creating leaderboards traditionally demands significant manual effort. In recent years, efforts have been made to automate leaderboard generation, but existing datasets for this purpose are limited by capturing only the best results from each paper and limited metadata. We present \textsc{MetaLead}, a fully human-annotated \textbf{ML leaderboard} dataset that captures all experimental results for \textit{result transparency} and contains extra \textbf{metadata}, such as the result \textit{experimental type}(baseline, proposed method, or variation of proposed method) for \textit{experiment-type guided comparisons}, and explicitly separates \textit{train} and \textit{test dataset} for \textit{cross-domain assessment}. This enriched structure makes \textsc{MetaLead} a powerful resource for more transparent and nuanced evaluations across ML research. \textsc{MetaLead} dataset and code repository: \url{https://github.com/RoelTim/metalead}
\end{abstract}


\section{Introduction}\label{sec:intro}

The volume of scientific literature has grown exponentially\footnote{The scale of this increase is well demonstrated by arXiv, with a consistent annual growth in publication volume of approximately 10\%~\cite{arxiv2024submissions}.}, resulting in increased \textit{cognitive load} for researchers, who struggle to compare existing work due to a \textit{scientific information overload}~\cite{landhuis2016scientific,bornmann2021growth}. This information overload for researchers, combined with the fast-paced development of AI, has led to a growing supply of \textit{AI for Science} tools designed to help researchers efficiently process and understand existing research, such as \textit{Google's Co Scientist}~\cite{gottweis2025towards}, \textit{Ought's Elicit}~\cite{elicit}, and, \textit{OpenAI's Deep Research}~\cite{openai2025deepresearch}. 

For data science workflows, we note that the field already employs various conventions to assist in research planning and evaluation.  For example, in Machine Learning (ML), a tool to easily compare the performance of different models is \textit{ML leaderboards}. A popular online platform for ML leaderboards is \textit{Papers with Code}\footnote{\url{https://paperswithcode.com/}} {\color{green}, which links research papers to their associated code implementations and maintains community-curated leaderboards across tasks.}. 

However, manually curating these types of leaderboards is labour-intensive and requires continual manual maintenance. To ease this burden and enable \textit{cognitive offloading}\footnote{Cognitive offloading refers to the use of external tools or strategies to reduce mental effort.}~\cite{risko2016cognitive}, researchers have developed NLP-based tools that automate the curation process~\cite{singh2019automated, Hou2019Leaderboards,Kardas2020AXCELL,Kabongo2021Mining,Yang2022TELIN,kabongo2024effective}. With the first publication in 2019, based on a BERT-style transformer encoder trained for natural language inference (NLI) \cite{Hou2019Leaderboards}, while recent publications are based on multi-agent LLM frameworks \cite{sahinuc2024efficient}.

\begin{figure}
    \centering
    \includegraphics[width=\linewidth]{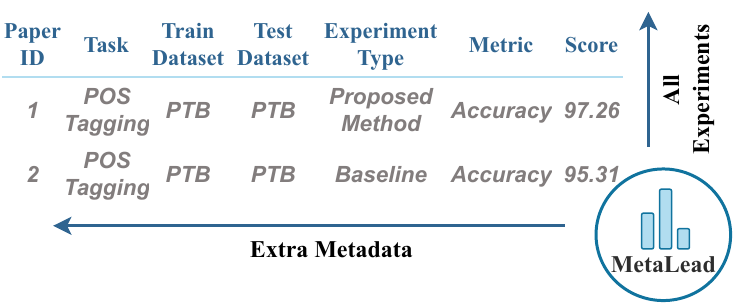}
    \caption{\textsc{MetaLead} is more comprehensive than existing leaderboard datasets, with broader metadata (horizontal) and full inclusion of experimental results (vertical).}
    \label{fig:metalead-intro}
\end{figure}

\begin{table*}[tbh]
\setlength{\tabcolsep}{4pt}
\centering
\footnotesize
\begin{tabularx}{\linewidth}{lclX}
\toprule \textbf{Dataset} & \textbf{All Results} & \textbf{Annotation Source}& \makecell[l]{\textbf{Scope Tuples}} \\ \toprule
\citet{Hou2019Leaderboards} & \xmark &NLP Progress&\textit{$\langle$task, dataset, metric, score$\rangle$} \\
\citet{Kardas2020AXCELL}& \xmark & Papers with Code &\textit{$\langle$task, dataset, metric, score$\rangle$} \\ 
\citet{kabongo2024effective} & \xmark &Papers with Code &\textit{$\langle$task, dataset, metric, score$\rangle$} \\ 
\citet{Singh2024LEGOBENCH} &\xmark&Papers with Code&\textit{$\langle$task, dataset, metric, method, score$\rangle$}\\
\citet{sahinuc2024efficient} &\xmark&Human Annotation& \textit{$\langle$task, dataset, metric, score$\rangle$}\\
\textsc{MetaLead} (Ours)& \cmark &Human Annotation&\textit{$\langle$task, train dataset, test dataset, metric, score, experiment type$\rangle$} \\ \bottomrule
\end{tabularx}
\caption{Comparison of \textsc{MetaLead} and related datasets. \textsc{MetaLead} uniquely includes all reported results, uses human annotation, and captures extended tuples with dataset splits and \textit{experiment type}. }
\label{tab:dataset_info}
\end{table*}

Even though NLP methods for extracting ML leaderboard entries have advanced, the scope of leaderboards remains limited~\cite{timmer-etal-2025-position}. Existing leaderboards (\autoref{sec:related-work}) report only the best results, omitting ablations, negative results, and baselines. For instance, a recent shared task extracted only \textit{$\langle$task, dataset, metric, score$\rangle$} tuples~\cite{d2024overview}, reflecting the field’s current focus while neglecting broader experimental context.

The limited scope of current leaderboards leads to two key \textit{shortcomings}. First, focusing only on best-performing results causes \textit{performance myopia}\footnote{Performance myopia is the tendency to focus solely on best-performing results, neglecting broader context such as negative results, baselines, or failed variants.
}~\cite{ethayarajh2020utility} and promotes publication bias, favouring positive results~\cite{karl2024position}. Second, the lack of distinction between train and test datasets hinders meaningful \textit{in-domain} (ID) versus \textit{out-of-domain} (OOD) comparisons, a critical distinction, as performance often varies across domains, leading to different conclusions and downstream consequences~\cite{teney2020value,gebru2021datasheets}.

To address these shortcomings, we introduce \textsc{MetaLead}, building on the previously annotated document set \textit{SciLead} introduced by \citet{sahinuc2024efficient}, {\color{green}a manually annotated benchmark of scientific leaderboards that captures (task, dataset, metric, score) tuples from ML papers.}  Our additional annotations are illustrated in \autoref{fig:metalead-intro}, adding 3,285 new records to the original 295 best records (i.e., leaderboard result) of \citet{sahinuc2024efficient}. Although \textsc{MetaLead} diverges from conventional leaderboards by including all reported results, it preserves the leaderboard’s core purpose: enabling comparative evaluation of model performance. This broader inclusion enhances transparency rather than undermining the concept.

The three main contributions of this work are as follows: First, we introduce \textbf{\textsc{MetaLead}}, the first leaderboard-style dataset to include \textit{all} experimental results from papers, enabling more transparency (\autoref{sec:result_transparancy}). Second, we expand the schema of leaderboard tuples in two ways: (1) by introducing an explicit \textit{experiment type} field, which enables experiment-type guided comparisons (\autoref{sec:experiment-type-guided-comparison}) and is accompanied by a novel metric for evaluation (\autoref{sec:experiment-type-evaluation});\, and (2) by adding an explicit \textit{train/test dataset distinction}, which supports cross-domain assessment (\autoref{sec:cross-domain-assessment}). Third, we provide a fully \textbf{human-annotated} resource (\autoref{sec:metalead-annotation}) and benchmarking results across various LLMs (\autoref{sec:evaluation}-\autoref{sec:experiments}). Overall, \textsc{MetaLead} represents a step towards more holistic ML evaluation, broadening the notion of leaderboards and enabling more rigorous and transparent scientific inquiry.

\section{Related Work}\label{sec:related-work}

\paragraph{ML Leaderboard Datasets}

\textsc{MetaLead} builds on prior work in ML leaderboard generation. ~\autoref{tab:dataset_info} summarises existing datasets and shows how \textsc{MetaLead} compares. The comparison includes only datasets that report experimental results, as leaderboards are only meaningful when scores are provided.\footnote{E.g., \citet{Kabongo2021Mining} report \textit{$\langle$task, dataset, metric$\rangle$} tuples but do not include scores.} A key distinction of \textsc{MetaLead} is its comprehensive coverage: \textsc{MetaLead} includes all experimental results, whereas prior datasets report only the best results. Most existing datasets rely on community annotations, {\color{green} for example Papers with Code and its precursor NLP Progress\footnote{\url{https://nlpprogress.com/}, a GitHub community-maintained NLP leaderboard.}} 
In contrast, only \textsc{MetaLead} and \citet{sahinuc2024efficient} provide fully human-annotated leaderboards, ensuring high-quality annotations. Tuple scope further differentiates datasets: most extract \textit{$\langle$task, dataset, metric, score$\rangle$} tuples, while \citet{Singh2024LEGOBENCH} reports \textit{$\langle$method, score$\rangle$} conditioned on \textit{$\langle$task, dataset, metric$\rangle$}. \textsc{MetaLead}'s structure is the most detailed: \textit{$\langle$task, train dataset, test dataset, metric, score, experiment type$\rangle$}, distinguishing training and test datasets and categorising results as baselines, proposed methods, or variations.

\paragraph{Scientific Information Extraction}
Automatic ML leaderboard generation is part of a wider body of work about scientific information extraction (IE). \citet{Jain2020SCIREX} introduced SciREX, a benchmark dataset for scientific relation extraction focusing on identifying experimental results and their context within scientific documents. Similarly, \citet{hou2021tdmsci} proposed TDM-Sci, which focuses on structured extraction of \textit{$\langle$task, dataset, metric$\rangle$} triples from scientific papers. Other examples are IE used for scientific knowledge graph construction~\cite{ammar2018construction, luan2018multi, mondal2021end,timmer2023KG,khoo2023KG} and the extraction of citations and entities~\cite{jurgens2018measuring}

\section{Schema Design Choices for \textsc{MetaLead}}\label{sec:context-for-leaderboards}

\begin{figure*}[tb]
    \centering
    \includegraphics[width=\textwidth]{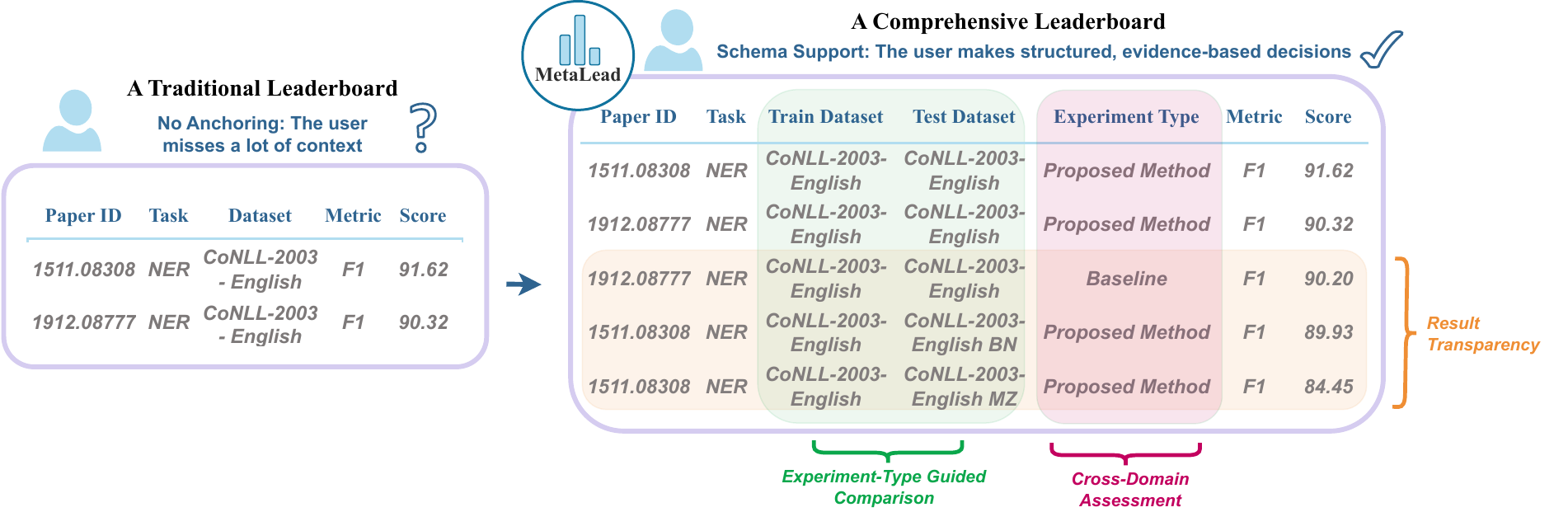}
    \caption{Schema benefits introduced by \textsc{MetaLead}. Compared to traditional leaderboards, \textsc{MetaLead} enables structured analysis through three key schema innovations: 1) \textit{result transparency} by including all reported outcomes, 2) \textit{experiment-type guided comparisons} via explicit labelling of baselines, proposed methods, and variations, and 3) \textit{cross-domain assessment} by distinguishing between training and test datasets.
    }
    \label{fig:before-after-metalead}
\end{figure*}

\textsc{MetaLead} is designed to extend the leaderboard format to serve as a \textit{research planning scaffold} for interpreting experimental results. \textsc{MetaLead} builds on existing human-curated leaderboard datasets by promoting \textit{result transparency} (\autoref{sec:result_transparancy}), \textit{experiment-type guided comparison} (\autoref{sec:experiment-type-guided-comparison}), and \textit{cross-domain assessment} (\autoref{sec:cross-domain-assessment}). Each design choice addresses specific bottlenecks in interpreting ML experiments at scale, which have been previously identified and discussed in the literature. Figure~\ref{fig:before-after-metalead} shows a diagram that visualises how these three design choices impact leaderboards. 

\subsection{Result Transparency}\label{sec:result_transparancy}

All existing leaderboard datasets include only the best-performing results, promoting \textit{performance myopia} and \textit{reinforcing confirmation bias}~\cite{ethayarajh2020utility}. Science suffers from publication bias, favouring positive results, and leaderboards exacerbate this by showcasing only top results~\cite{karl2024position}.  {\color{green}Previous ML generation systematically point out that a key direction for future research is including all experimental results, for example \citet{timmer-etal-2025-position} advocate for vertical expansion of leaderboards, \citet{Hou2019Leaderboards} call it an ideal situation to extract all results reported in the same paper, \citet{Kardas2020AXCELL} point out that retrieving all results rather than just the principal results introduced in the paper would enable fine-grained comparisons.}

{\color{green}Reporting bias is a big problem in the ML field, for example \citet{mcgreivy2024weak} found that 79\% of papers compare only against weak baselines. Including all baselines in leaderboards, such as in \textsc{MetaLead}, increases transparency and reduces the risk that researchers inadvertently or deliberately select weak baselines.}

In contrast, the \textsc{MetaLead} schema captures \emph{all} reported results, including ablations, negative outcomes, and failed variations. This component supports what we term \emph{result transparency}: the idea that decision quality improves when users are exposed to all the experimental results rather than a curated success narrative.

\subsection{Experiment-Type Guided Comparisons}\label{sec:experiment-type-guided-comparison}
Including all experimental results can overwhelm users if not paired with an appropriate experimental description. {\color{green}Previous work vow for enriching leaderboard by providing more experimental type information, for example, \citet{kabongo2024orkg} are interested in extend the current triples (task, dataset, metric) model with additional concepts, \citet{timmer-etal-2025-position} recommend expanding leaderboards horizontally, and \citet{Kardas2020AXCELL} mentions future work may want to build on our approach for more comprehensive extraction task. However, none of these works is detailed about what exactly this expansion should entail.} To facilitate meaningful comparisons, the \textsc{MetaLead} schema classifies each result by its \emph{experiment type}: \textit{baseline}, \textit{proposed method}, or \textit{variation of proposed method}. This categorical distinction provides \textit{experimental context} that helps researchers organise the landscape of results and interpret empirical evidence within its proper context.  By explicitly encoding the role of each result, the schema reduces ambiguity, supports faster categorisation and enables faster insight.

\subsection{Cross-Domain Assessment}\label{sec:cross-domain-assessment}
Existing leaderboard datasets' schemas force the extraction of a single dataset mention, introducing an ambiguity regarding whether the mention pertains to the training and evaluation contexts.
This obscures whether the evaluation results are In-Domain (ID) or Out-of-Domain (OOD). Prior research has shown that ID and OOD evaluations can yield markedly different results \cite{teney2020value, gebru2021datasheets}. \citet{liu2021explainaboard} of ExplainaBoard, advocates a partitioning of the test set, allowing researchers to observe where models perform well or poorly and to gain insights beyond overall accuracy scores. The \textsc{MetaLead} schema addresses this by explicitly separating the \textit{train} and \textit{test} datasets. {\color{green}By listing the train and test sets separately in \textsc{MetaLead}, researchers can more easily identify patterns, e.g., the agreement-on-the-line described by \citet{baek2022agreement}, a distributional robustness phenomenon regarding ID and OOD in ML, which we describe as \textit{cross-domain assessment}.} 
\section{\textsc{MetaLead} Dataset Annotation}\label{sec:metalead-annotation}

\autoref{tab:metalead-scilead-stats} presents the statistics of the fully human-annotated \textsc{MetaLead} dataset, constructed through a two-step annotation process: step 1: core tuple annotation (\autoref{sec:core-tuple-annotation}) and step 2: experiment type labelling (\autoref{sec:experiment-type-labelling}). We outline key annotation challenges encountered during the process (\autoref{sec:annotation-challenges}), and describe how the leaderboards are constructed from the annotated tuples (\autoref{sec:leaderboard-construction}).


\begin{table}[t]
\setlength{\tabcolsep}{4pt}
\footnotesize
\begin{tabularx}{\linewidth}{l l X}
\toprule
 & \multicolumn{2}{c}{\textbf{Count}} \\ 
\cmidrule(lr){2-3} 
\textbf{Dataset Item}& \textbf{SciLead} & \textbf{\textsc{MetaLead}} \\ 
\midrule
\textbf{General Statistics}\\
\quad Papers & 43& 43 \\
\quad Total tuples & 295&3,568\\
\quad Avg. Annotations per Paper & 7 &83 \\
\quad Median Annotations per Paper & 6&61 \\
\textbf{Diversity of Content}\\
\quad Unique Leaderboards& 27 & 283\\
\quad Unique Tasks & 23&23 \\
\quad Unique Train Datasets & - &83\\
\quad Unique (Test) Datasets & 71 &160\\
\quad Unique Metrics &26 &26\\
\textit{Experimental Type}\\
\quad Baseline & - & 1,919\\
\quad Proposed Method & - & 809\\
\quad Variation of Proposed Method & - & 840\\
\bottomrule
\end{tabularx}
\caption{Statistics of our \textsc{MetaLead} dataset versus SciLead. \textbf{-} if not defined, as these are not existing categories/labels in the SciLead dataset. \label{tab:statistics}}
\label{tab:metalead-scilead-stats}
\end{table}

\begin{figure}[bh]
    \centering
    \includegraphics[width=.9\linewidth]{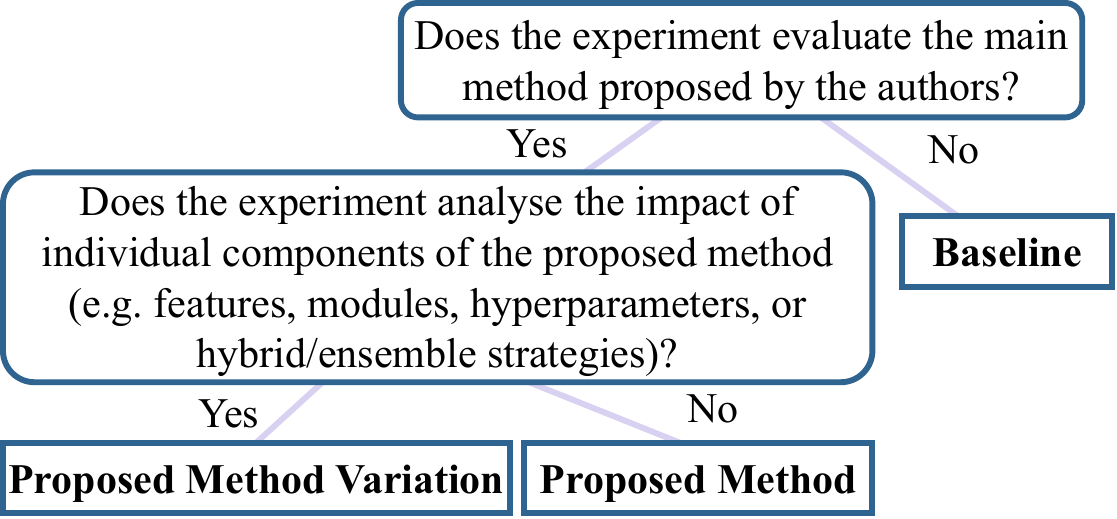}
    \caption{Decision tree for the \textit{experiment type} entry of \textsc{MetaLead}.}
    \label{fig:experiment-type}
    
\end{figure}

\subsection{Core Tuple Annotation}\label{sec:core-tuple-annotation}
The annotation process was carried out in two distinct steps. The annotators are the authors of this paper, further details are provided in the Ethics section. Step 1 is annotating the core tuple: \textit{$\langle$task, train dataset, test dataset, metric, score$\rangle$}. In line with the quality-control single-annotator approach introduced by \citet{sahinuc2024efficient}, the annotator annotates all the experimental results reported in the paper. The annotation instructions for step 1 are provided in \autoref{app:annotation_guidelines} ~\autoref{fig:instructions_annotations1}.

\subsection{Experiment Type Labelling}\label{sec:experiment-type-labelling}

Step 2 is labelling the \textit{experiment type}. \autoref{fig:experiment-type} shows the decision tree for annotating \textit{experiment type} and the detailed annotation instructions are provided in \autoref{app:annotation_guidelines} ~\autoref{fig:instructions_annotations2}. To ensure clarity and consistency for this \textit{potentially} more subjective classification task, two annotators independently annotated an initial batch of five randomly selected papers, we observed a Cohen's Kappa of 0.91, indicating \textit{near-perfect agreement}~\cite{cohen1960coefficient}. We then held a resolution discussion to address the disagreements, during which we revised both the annotation guidelines and the corresponding annotations, resulting in perfect agreement (Cohen's Kappa = 1.00). Using the updated guidelines, we annotated a second batch of five papers and again observed near-perfect agreement (Cohen's Kappa = 0.94). Given the consistently high inter-annotator agreement, the remaining papers were annotated by a single annotator using the finalised guidelines.

\subsection{Annotation Challenges}\label{sec:annotation-challenges}

Annotation of papers for leaderboards presents several challenges. A common issue is that authors are not explicit about whether they propose a single main method or multiple methods. For example, \citet{narayan2018don} propose four variants of a \textit{T-CONVS2S} architecture, but the authors do not make clear whether one variant is the main method or whether all should be considered equivalent. An example of inconsistency within a paper appears in the work by \citet{li2019dual}, where the authors describe a method as a baseline in the main text, yet label it as ``Ours'' rather than ``Baseline'' in the corresponding results table. These inconsistencies complicate the annotation process and introduce ambiguity in classifying experimental results.

\subsection{Leaderboard Construction}\label{sec:leaderboard-construction}
We construct a leaderboard for each unique combination of \textit{$\langle$task, train dataset, test dataset, metric, score$\rangle$}, resulting in a total of 283 leaderboards. Whereas some existing leaderboard datasets impose a minimum threshold to consider a leaderboard valid (e.g., \citet{Kabongo2021Mining} require five papers per leaderboard), we do not apply such thresholds. The aim of \textsc{MetaLead} is to offer a more granular view of experimental results. Imposing a threshold would have excluded important information, particularly for settings such as ID versus OOD evaluation.
\section{Evaluation}\label{sec:evaluation}

We evaluate extraction performance in open and closed domains (\autoref{sec:domain_evaluation}) using tuple- and entity-level metrics (\autoref{sec:entity_evaluation}), propose a new metric for \textit{experiment type} classification accuracy (\autoref{sec:experiment-type-evaluation}), and introduce leaderboard-specific metrics to assess broader utility (\autoref{sec:leaderboard_evaluation}).

\subsection{Open and Closed Domain Scenario}\label{sec:domain_evaluation}
Following \citet{timmer-etal-2025-position}, we evaluate both open and closed domain settings.\footnote{Also known as the \textit{cold-start} scenario~\cite{sahinuc2024efficient}.} In the open domain, the model receives no candidate tuples and must extract all relevant entries directly from the paper. In the closed domain, the model selects entities from a predefined list. For open domain evaluation, we apply paper-level normalisation, as described by \citet{sahinuc2024efficient}.

\subsection{Extraction Level Metrics}\label{sec:entity_evaluation}

Given the novel nature of a task for comprehensive extraction of results, we introduce and define our evaluation metrics here.  We report the \textit{set-based} recall, precision, and F1, \textit{averaged across all papers}. Recall (\textbf{R}) is given by:

\vspace{-1em}\begin{equation}
    \text{\textbf{R}} = \frac{\sum_{p\in P} \frac{|E_p^G \cap E_p^M  | }{|E_p^G|} }{P}
\end{equation}\vspace{-1em}

\noindent for all papers $P$, recall is computed per paper $p \in P$ as the ratio of correctly predicted entities (or tuples), $|E_p^G \cap E_p^M|$, to the total number of gold entities, $|E_p^G|$. Here, $|\cdot|$ denotes the \textit{cardinality} of a set, i.e., the number of unique elements. The overall recall is the average recall across all papers.

Precision (\textbf{P}) is given by:

\vspace{-1em}\begin{equation}
\text{\textbf{P}} = \frac{\sum_{p\in P} \frac{|E_p^G \cap E_p^M | }{|E_p^M|} }{P}
\end{equation}\vspace{-1em}

\noindent with for each paper $p \in P$, precision is computed as the ratio of correctly predicted entities $|E_p^G \cap E_p^M|$ to the total number of predicted entities $|E_p^M|$. 

Based on \textbf{R} and \textbf{P}, we also report the \textbf{F1} score. Together, the set-based \textbf{R}, \textbf{P}, and \textbf{F1} metrics assess how well the model captures relevant entities across the corpus, while avoiding overcounting due to repeated predictions. They offer a more conservative measure of performance by focusing on the presence of correct entities rather than frequency.

\subsection{Experiment Type Evaluation}\label{sec:experiment-type-evaluation}

We propose a novel metric to evaluate \textit{experiment type} classification: for each correctly extracted \textit{$\langle$task, train dataset, test dataset, metric, score$\rangle$} tuple, we compute the percentage with correctly assigned \textit{experiment type}, \textbf{ET-Accuracy}:

\vspace{-1em}
\begin{equation}
\text{\textbf{ET-Accuracy}} = \frac{\sum_{p \in P} \frac{|\hat{T}_p \cap T_p|}{|\hat{T}_p|}}{P}
\end{equation}
\vspace{-1em}

\noindent where $\hat{T}_p$ is the set of predicted tuples for paper $p$ with a correct core tuple, and $T_p \subseteq \hat{T}_p$ contains those with a correct \textit{experiment type} label. To better understand classification difficulty, we compute confusion matrices. Each confusion matrix is normalised per true label (i.e., each row sums to 100\%), and we report the average across all non-empty matrices, where rows with no true instances are excluded from the average.

\subsection{Leaderboard Level Metrics}\label{sec:leaderboard_evaluation}

Following \citet{sahinuc2024efficient}, we calculate four leaderboard-specific metrics: \textit{Leaderboard Recall}, \textit{Paper Coverage}, \textit{Result Coverage}, and, \textit{Average Overlap} .

We define $\mathcal{L}$ as the set of gold leaderboards. For each leaderboard $\ell \in \mathcal{L}$, let $P^{\text{gold}}_\ell$ and $P^{\text{pred}}_\ell$ be the sets of gold and predicted papers. Leaderboard Recall (\textbf{LR}) is calculated by seeing how many of the leaderboards have been created, \textit{i.e.,} how many of the unique golden \textit{$\langle$task, train dataset, test dataset, metric, experiment type$\rangle$} tuples have been reported in the predicted:

\vspace{-1em}\begin{equation}
\text{\textbf{LR}} = \frac{| \mathcal{L}^{\text{gold}} \cap \mathcal{L}^{\text{pred}} |}{| \mathcal{L}^{\text{gold}} |}
\label{eq:lr}
\end{equation}\vspace{-1em}

Paper Coverage (\textbf{PC}) is the average recall of assigned papers\footnote{We treat paper assignments as sets and ignore duplicate entries per leaderboard.}:

\vspace{-1em}\begin{equation}
\text{\textbf{PC}} = \frac{1}{|\mathcal{L}|} \sum_{\ell \in \mathcal{L}} \frac{|P^{\text{gold}}_\ell \cap P^{\text{pred}}_\ell|}{|P^{\text{gold}}_\ell|}
\end{equation}\vspace{-1em}

Result Coverage (\textbf{RC}) is computed analogously, where $R^{\text{gold}}_\ell$ denotes the set of ground truth results on leaderboard $\ell$, and $R^{\text{pred}}_\ell$ denotes the set of predicted results on leaderboard $\ell$:

\vspace{-1em}\begin{equation}
\text{\textbf{RC}} = \frac{1}{|\mathcal{L}|} \sum_{\ell \in \mathcal{L}} \frac{|R^{\text{gold}}_\ell \cap R^{\text{pred}}_\ell|}{|R^{\text{gold}}_\ell|}
\end{equation}\vspace{-1em}

Average Overlap (\textbf{AO}) \cite{webber2010similarity} is a metric to measure the similarity between two ranked leaderboards:

\vspace{-1em}\begin{equation}
\text{\textbf{AO}} = \frac{1}{|\mathcal{L}|} \sum_{\ell \in \mathcal{L}} \frac{1}{d} \sum_{i=1}^{d} \frac{|r^{\text{gold}}_{\ell,1:i} \cap r^{\text{pred}}_{\ell,1:i}|}{i}
\label{eq:ao}
\end{equation}\vspace{-1em}

\noindent where, $r^{\text{gold}}_{\ell,1:i}$ and $r^{\text{pred}}_{\ell,1:i}$ denote the top-$i$ elements of the gold and predicted ranks of leaderboard $\ell$.

\section{Experiments}\label{sec:experiments}

\paragraph{Experimental Setup} 

We conduct experiments using proprietary, open source\footnote{We use the term open source, although in practice the openness may be limited, e.g. only weights may be released.}, and baseline models. Proprietary models include GPT-4.1~\cite{openai2024gpt41}, GPT-4o~\cite{openai2024gpt4o_system_card}, o4-mini~\cite{openai2025o3o4mini}, Gemini-2.5-Pro, Gemini-2.5-Flash~\cite{comanici2025gemini}, and Gemini-1.5-Pro~\cite{team2024gemini}. Open source models include Llama 3.3 70B Instruct~\cite{dubey2024llama} and Mistral Large~\cite{mistral-huggingface-mistral-large}. Prompts are shown in \autoref{sec:appendix_prompts}, and more experimental details in \autoref{sec:appendix_exp_setup_details}. We implement modified versions of the baselines by \citet{sahinuc2024efficient}\footnote{\url{https://github.com/UKPLab/arxiv2024-leaderboard-generation}} and \citet{Kardas2020AXCELL}\footnote{\url{https://github.com/paperswithcode/axcell}}. For \citeauthor{sahinuc2024efficient}, we updated the prompt to match the new schema. For \citeauthor{Kardas2020AXCELL}, we extended the code to extract all results, though the output remains limited to $\langle$task, dataset, metric, score$\rangle$ tuples. Where possible, we replaced the original LLMs of these baselines with GPT-4.1 for a fairer comparison.

\paragraph{Tuple Extraction Results}

\autoref{tab:exact_tuple_match} shows exact tuple match results for \textit{$\langle$task, train dataset, test dataset, metric, score, experiment type$\rangle$} in both closed- and open-domain settings. As expected, closed-domain results are higher, though this setting is less realistic. GPT-4.1 and Gemini-2.5-Pro outperform other models. Gemini-1.5-Pro shows high precision but low recall, indicating correct but incomplete extraction. All proprietary models outperform open source models, and all LLMs outperform the baselines~\cite{sahinuc2024efficient, Kardas2020AXCELL}. The baselines underperform due to their original design for a different schema and reliance on text filtering strategies, which were necessary for earlier LLMs with smaller context windows. In contrast, current models benefit from full-text processing.

\begin{table}[tbh]
    \setlength{\tabcolsep}{3pt}
    \footnotesize
    \centering 
    \footnotesize
    \begin{tabular}{lll r r r}
        \toprule
        & \textbf{Type} & \textbf{Name} & \textbf{P} & \textbf{R} & \textbf{F1} \\
        \midrule
        \multirow{10}{*}{\rotatebox[origin=c]{90}{\textit{Closed Domain}}}&\multirow{5}{*}{\textbf{Proprietary}}&GPT-4.1&46.04&38.16&40.21\\
        &&GPT-4o &38.65&9.76&13.93\\
        &&o4-mini&45.88&36.79&39.42\\
        &&Gemini-2.5-Pro&{\color{customblue}\textbf{46.55}}&{\color{customblue}\textbf{52.34}}&{\color{customblue}\textbf{48.89}}\\
        &&Gemini-2.5-Flash&17.78&19.13&18.19\\
        &&Gemini-1.5-Pro&42.47&17.42&21.98\\
        \arrayrulecolor{gray!50}\cmidrule(lr){2-6}
    \arrayrulecolor{black}
        &\multirow{2}{*}{\textbf{\makecell[l]{Open\\Source}}}&Llama 3.3 70B &37.42&24.82&26.96\\
        &&Mistral Large&40.43&24.86&28.21\\
        \arrayrulecolor{gray!50}\cmidrule(lr){2-6}
    \arrayrulecolor{black}
    &\multirow{2}{*}{\textbf{Baselines}}&\citet{sahinuc2024efficient}&18.41&17.78&18.10\\        &&\citet{Kardas2020AXCELL}\textsuperscript{*}&14.45&79.69&76.37\\
        \midrule
        \multirow{8}{*}{\rotatebox[origin=c]{90}{\textit{Open Domain}}}&\multirow{5}{*}{\textbf{Proprietary}}&GPT-4.1&16.01&13.49&14.14\\
        &&GPT-4o&33.50&9.14&12.61\\
        &&o4-mini&37.23&{\color{customorange}\textbf{28.02}}&{\color{customorange}\textbf{29.99}}\\
        &&Gemini-2.5-Pro&24.32&27.02&25.48\\
        &&Gemini-2.5-Flash&16.65&17.61&16.81\\
        &&Gemini-1.5-Pro&16.54&8.38&9.90\\
        \arrayrulecolor{gray!50}\cmidrule(lr){2-6}
    \arrayrulecolor{black}
        &\multirow{2}{*}{\textbf{\makecell[l]{Open\\Source}}}&Llama 3.3 70B&{\color{customorange}\textbf{37.51}}&23.14&25.78\\
        &&Mistral Large&33.77&24.02&26.60\\
        \arrayrulecolor{gray!50}\cmidrule(lr){2-6}
    \arrayrulecolor{black}
        &\textbf{Baselines}&\citet{sahinuc2024efficient}&1.91&2.14&2.02\\
        \bottomrule
        \multicolumn{6}{l}{
  \parbox[t]{\linewidth}{\textsuperscript{*} The scope of the tuple for ~\cite{Kardas2020AXCELL} is limited to: \textit{$\langle$task, (test) dataset, metric, score$\rangle$}, and therefore these scores are not considered to be top results.}}
    \end{tabular}
    \caption{Exact Tuple Match for \textsc{MetaLead} for \textit{$\langle$task, train dataset, test dataset, metric, score, experiment type$\rangle$}, with top results in the closed domain highlighted in {\color{customblue}\textbf{blue}} and in the open domain in {\color{customorange}\textbf{orange}}.}
    \label{tab:exact_tuple_match}
\end{table}

\begin{table*}[tbh]
    \setlength{\tabcolsep}{3pt}
    \centering 
    \footnotesize
    \resizebox{\linewidth}{!}{
    \begin{tabular}{lll ccc ccc ccc ccc ccc}
    \toprule
    &\multicolumn{2}{c}{\textbf{Model}}&\multicolumn{3}{c}{\textbf{Task}} & 
    \multicolumn{3}{c}{\textbf{Train Dataset}} & 
    \multicolumn{3}{c}{\textbf{Test Dataset}} & 
    \multicolumn{3}{c}{\textbf{Metric}} & 
    \multicolumn{3}{c}{\textbf{Result}} \\
    \cmidrule(lr){2-3}
    \cmidrule(lr){4-6} \cmidrule(lr){7-9} \cmidrule(lr){10-12}
    \cmidrule(lr){13-15} \cmidrule(lr){16-18}
    &\textbf{Type}&\textbf{Name}& P & R & F1 & P & R & F1 & P & R & F1 & P & R & F1 & P & R & F1 \\
    \midrule
     \multirow{10}{*}{\rotatebox[origin=c]{90}{\textit{Closed Domain}}}&\textbf{Proprietary} & GPT-4.1 & \color{customblue}\textbf{97.67} & \color{customblue}\textbf{99.77} & {\color{customblue}\textbf{98.33}} & 81.82 & 78.14 & 78.86 & 81.42 & 71.89 & 74.02 & 88.99 & 82.62 & 83.93 & 93.93 & 67.22 & 74.16 \\
    && GPT-4o & 95.40 & 97.03 & 95.28 & 80.84 & 72.96 & 75.00 & 76.58 & 62.04 & 65.96 & 89.26 & 72.65 & 77.77 & 95.23 & 24.11 & 34.48 \\
    && o4-mini & 89.40 & 97.44 & 91.71 & 86.65 & 81.98 & {\color{customblue}\textbf{83.06}} & 76.72 & 65.66 & 68.36 & 86.07 & 82.47 & 82.33 & 96.82 & 72.02 & 77.82 \\
    && Gemini-2.5-Pro & 90.11 & 99.51 & 92.85 & 81.15 & {\color{customblue}\textbf{83.33}} & 81.58 & 81.88 & {\color{customblue}\textbf{83.42}} & {\color{customblue}\textbf{81.59}} & 78.11 & {\color{customblue}\textbf{88.01}} & 81.32 & 88.40 & {\color{customblue}\textbf{99.07}} & {\color{customblue}\textbf{92.74}} \\
    && Gemini-2.5-Flash & 73.47 & 85.47 & 77.40 & 59.98 & 68.92 & 62.33 & 56.38 & 58.32 & 55.83 & 66.59 & 73.25 & 68.18 & 89.16 & 95.92 & 91.28 \\
    && Gemini-1.5-Pro & 93.02 & 96.12 & 92.87 & 79.21 & 66.11 & 69.62 & 77.61 & 57.91 & 62.78 & {\color{customblue}\textbf{91.99}} & 78.37 & 82.98 & {\color{customblue}\textbf{97.92}} & 34.67 & 45.62 \\
    \arrayrulecolor{gray!50}\cmidrule(lr){2-18}
    \arrayrulecolor{black}
    &\multirow{2}{*}{\makecell[l]{\textbf{Open}\\ \textbf{Source}}} & Llama 3.3 70B & 89.65 & 96.98 & 91.82 & 76.02 & 68.84 & 70.71 & 76.22 & 61.19 & 64.66 & 83.62 & 73.43 & 75.58 & 95.67 & 49.71 & 58.70 \\
    && Mistral Large  & 93.60 & 96.01 & 92.66 & {\color{customblue}\textbf{88.07}} & 77.15 & 80.21 & {\color{customblue}\textbf{86.60}} & 64.80 & 70.24 & 80.97 & 73.96 & 75.40 & 93.65 & 47.27 & 56.80 \\
    \arrayrulecolor{gray!50}\cmidrule(lr){2-18}
    \arrayrulecolor{black}
    &\multirow{2}{*}{\textbf{Baselines}} 
    &\citet{sahinuc2024efficient}& 88.69 & 97.41 & 92.85 & 75.17 & 75.60 & 75.38 & 43.57 & 36.65 & 39.81 & 87.42 & 81.19 & {\color{customblue}\textbf{84.19}} & 90.84 & 70.22 & 79.21 \\
    &&\citet{Kardas2020AXCELL} & 78.06&79.69&76.37&-&-&-&56.62&44.89&46.65&50.91&66.10&53.20&74.40&83.24&77.68\\
    \midrule
    \multirow{9}{*}{\rotatebox[origin=c]{90}{\textit{Open Domain}}}&\textbf{Proprietary} & GPT-4.1 & 57.75 & 53.47 & 54.23 & 70.60 & 61.91 & 64.26 & 65.11 & 49.79 & 53.86 & 83.60 & 76.94 & 78.74 & 93.95 & 71.19 & 77.13 \\
    && GPT-4o & 90.70 & 88.14 & 87.67 & 76.51 & 65.64 & 69.44 & 78.29 & 56.23 & 62.04 & {\color{customorange}\textbf{90.47}} & 72.95 & 78.29 & 97.24 & 24.06 & 34.99 \\
    && o4-mini & 89.53 & 96.47 & {\color{customorange}\textbf{91.19}} & 78.54 & {\color{customorange}\textbf{72.89}} & {\color{customorange}\textbf{73.94}} & 76.68 & 58.69 & 62.87 & 80.51 & 75.68 & 75.90 & 94.93 & 71.46 & 76.41 \\
    && Gemini-2.5-Pro & 77.56 & 80.45 & 76.56 & 72.06 & 67.48 & 68.38 & 69.50 & 63.83 & 64.62 & 64.83 & 66.14 & 63.78 & 87.39 & {\color{customorange}\textbf{98.30}} & {\color{customorange}\textbf{91.87}} \\
    && Gemini-2.5-Flash & 85.99 & {\color{customorange}\textbf{98.50}} & 89.84 & 48.80 & 56.37 & 50.59 & 46.68 & 43.76 & 44.10 & 74.21 & {\color{customorange}\textbf{78.72}} & 75.43 & 88.04 & 93.33 & 89.26 \\
    && Gemini-1.5-Pro & 78.57 & 83.42 & 78.59 & 28.91 & 20.76 & 22.75 & 29.07 & 15.97 & 18.51 & 90.23 & 75.35 & {\color{customorange}\textbf{80.01}} & {\color{customorange}\textbf{98.19}} & 33.63 & 44.66 \\
    \arrayrulecolor{gray!50}\cmidrule(lr){2-18}
    \arrayrulecolor{black}
    &\multirow{2}{*}{\makecell[l]{\textbf{Open}\\\textbf{Source}}} & Llama 3.3 70B & 88.93 & 95.91 & 90.64 & 75.77 & 67.03 & 69.43 & {\color{customorange}\textbf{82.98}} & {\color{customorange}\textbf{64.80}} & {\color{customorange}\textbf{69.35}} & 87.27 & 75.22 & 78.64 & 92.62 & 47.22 & 56.32 \\
    && Mistral Large & {\color{customorange}\textbf{93.22}} & 89.22 & 88.72 & {\color{customorange}\textbf{84.14}} & 68.20 & 72.87 & 78.96 & 58.99 & 63.73 & 83.10 & 74.01 & 76.83 & 95.11 & 50.65 & 60.50 \\
    \arrayrulecolor{gray!50}\cmidrule(lr){2-18}
    \arrayrulecolor{black}
    &\textbf{Baselines} & \citet{sahinuc2024efficient} & 13.82 & 17.51 & 15.49 & 21.11 & 24.31 & 22.84 & 25.27 & 25.04 & 25.12 & 61.79 & 59.68 & 60.73 & 90.84 & 70.22 & 79.32 \\
    \bottomrule
    \end{tabular}
    }
    \caption{Model performance on individual entities in the \textsc{MetaLead} dataset for \textbf{open} and \textbf{closed} domains, with top closed-domain results in {\color{customblue}\textbf{blue}} and open-domain results in {\color{customorange}\textbf{orange}}.}
    \label{tab:IIM}
\end{table*}

\begin{table}[tbh]
    \setlength{\tabcolsep}{2pt}
    \centering 
    \footnotesize
    \resizebox{\linewidth}{!}{
    \begin{tabular}{lllrrrr}
    \toprule
        &\multicolumn{2}{c}{\textbf{Model}}\\
        \cmidrule{2-3}
         &\textbf{Type}&\textbf{Name}& \textbf{LR} & \textbf{PC} & \textbf{RC} & \textbf{AO}\\
         \hline
         \multirow{10}{*}{\rotatebox[origin=c]{90}{\textit{Closed Domain}}}&\textbf{Proprietary} & GPT-4.1&36.10&27.57&20.32&6.61\\
         &&GPT-4o&20.43&12.95&5.83&3.03\\
         &&o4-mini&30.11&22.80&14.97&5.95\\
         &&Gemini-2.5-Pro&{\color{customblue}\textbf{49.46}}&{\color{customblue}\textbf{41.05}}&{\color{customblue}\textbf{36.27}}&{\color{customblue}\textbf{7.99}}\\
         &&Gemini-2.5-Flash&23.96&13.98&11.79&{\color{customblue}\textbf{7.99}}\\
         &&Gemini-1.5-Pro&16.44&10.82&6.07&2.76\\
\arrayrulecolor{gray!50}\cmidrule(lr){2-7}
    \arrayrulecolor{black}
         &\multirow{2}{*}{\makecell[l]{\textbf{Open} \\ \textbf{Weigths}}} &
         Llama 3.3 70B&23.81&16.34&10.42&3.38\\
         &&Mistral Large&23.66&16.98&8.78&3.46\\
         \arrayrulecolor{gray!50}\cmidrule(lr){2-7}
    \arrayrulecolor{black}
         &\multirow{2}{*}{\textbf{Baselines}}&
         \citet{sahinuc2024efficient}&18.59&11.96&6.24&1.84 \\
         &&\citet{Kardas2020AXCELL}&14.50&6.99&5.95&0.00 \\
        \midrule
        \multirow{9}{*}{\rotatebox[origin=c]{90}{\textit{Open Domain}}}&\textbf{Proprietary} &GPT-4.1&7.07&5.47&3.42&0.92\\
         &&GPT-4o&13.67&8.74&4.43&2.00\\
         &&o4-mini & {\color{customorange}\textbf{25.19}} & {\color{customorange}\textbf{17.95}} & {\color{customorange}\textbf{12.80}}&{\color{customorange}\textbf{3.53}}\\
         &&Gemini-2.5-Pro&19.20&12.04&11.05&0.77\\
         &&Gemini-2.5-Flash&21.04&13.83&10.88&1.69\\
         &&Gemini-1.5-Pro&3.38&2.11&1.14&0.15\\
\arrayrulecolor{gray!50}\cmidrule(lr){2-7}
    \arrayrulecolor{black}
         &\multirow{2}{*}{\makecell[l]{\textbf{Open} \\ \textbf{Source}}} &Llama 3.3 70B&21.66&15.07&9.20&2.61\\
         &&Mistral Large&3.38&2.11&1.14&3.00\\
         \arrayrulecolor{gray!50}\cmidrule(lr){2-7}
    \arrayrulecolor{black}
         &\textbf{Baselines}&
         \citet{sahinuc2024efficient}&3.07&1.59&0.84&0.15\\
    \bottomrule
    \end{tabular}
    }
    \caption{Leaderboard Construction Results with leaderboard recall (LR), paper coverage (PC), results coverage (RC), and, average overlap (AO), with top closed-domain results in {\color{customblue}\textbf{blue}} and open-domain results in {\color{customorange}\textbf{orange}}.}
    \label{tab:leaderboard_results}
\end{table}

\begin{table}[tbh]
    \setlength{\tabcolsep}{1pt}
    \footnotesize
    \centering 
    \footnotesize
    \begin{tabular}{lllc}
    \toprule
        &\multicolumn{2}{c}{\textbf{Model}}\\
        \cmidrule{2-3}
         &\textbf{Type}&\textbf{Name}&\textbf{ET-Accuracy}\\
         \hline
         \multirow{10}{*}{\rotatebox[origin=c]{90}{\textit{Closed Domain}}}&\textbf{Proprietary} & GPT-4.1 & 70.88\\
         &&GPT-4o&58.13\\
         &&o4-mini&68.31\\
         &&Gemini-2.5-Pro&72.78\\
         &&Gemini-2.5-Flash&47.05\\
         &&Gemini-1.5-Pro&59.93\\
\arrayrulecolor{gray!50}\cmidrule(lr){2-4}
    \arrayrulecolor{black}
         &\multirow{2}{*}{\makecell[l]{\textbf{Open} \\ \textbf{Source}}} &
         Llama 3.3 70B&56.23\\
         &&Mistral Large&64.17\\
         \arrayrulecolor{gray!50}\cmidrule(lr){2-4}
    \arrayrulecolor{black}
         &\textbf{Baselines}&
         \citet{sahinuc2024efficient}&{\color{customblue}\textbf{81.62}}\\
        \midrule
        \multirow{10}{*}{\rotatebox[origin=c]{90}{\textit{Open Domain}}}&\textbf{Proprietary}&GPT4.1& 31.46\\
        &&GPT-4o&62.53\\
        &&o4-mini&63.72\\
        &&Gemini-2.5-Pro&53.27\\
        &&Gemini-2.5-Flash&50.24\\
        &&Gemini-1.5-Pro&33.85\\
\arrayrulecolor{gray!50}\cmidrule(lr){2-4}
    \arrayrulecolor{black}
         &\multirow{2}{*}{\makecell[l]{\textbf{Open} \\ \textbf{Source}}} &
         Llama 3.3 70B&59.87\\
         &&Mistral Large&62.96\\
         \arrayrulecolor{gray!50}\cmidrule(lr){2-4}
    \arrayrulecolor{black}
         &\textbf{Baselines}&
         \citet{sahinuc2024efficient}&{\color{customorange}\textbf{81.25}}\\
    \bottomrule
    \end{tabular}
    \caption{\textit{Experiment type} classification accuracy (\textbf{ET-Accuracy}), with top closed-domain results in {\color{customblue}\textbf{blue}} and open-domain results in {\color{customorange}\textbf{orange}}.}
    \label{tab:results_experiment-type}
\end{table}

\begin{table}[tbh]
\setlength{\tabcolsep}{2pt}
\centering
\footnotesize
\begin{tabular}{lrrr}
\toprule
\textbf{Gold / Predicted} & \textbf{Baseline} & \makecell[l]{\textbf{Proposed}\\\textbf{Method}} & \makecell[l]{\textbf{Method}\\\textbf{Variation}} \\
\midrule
\textbf{Baseline} & 89.67 & 0.66 & 9.68 \\
\textbf{Proposed Method} & 3.30 & 73.13 & 23.57 \\
\textbf{Method Variation} & 11.39 & 11.55 & 77.06 \\
\bottomrule
\end{tabular}
\caption{Confusion matrix for \textit{experiment type} classification by GPT-4.1 in the closed domain. Percentages are averaged over 39 papers with non-empty matrices.}
\label{tab:confusion-matrix}
\end{table}

\paragraph{Individual Entities Results}
\autoref{tab:IIM} present the results for the extraction of individual entities in the closed and open domain settings. The performance on individual entities is substantially higher than on complete tuples, as shown in \autoref{tab:exact_tuple_match}\footnote{Results for \textit{experiment type} are reported separately in Section~\autoref{tab:results_experiment-type} using a specialised metric.}. This indicates that extracting individual entities is considerably easier than correctly extracting their combinations. Consistent with the tuple extraction results, GPT-4.1 and Gemini-2.5-Pro achieve the highest scores for individual entity extraction. The lowest scores are generally observed for train  and test dataset, likely because the models must distinguish between them. The models also use varying terminology, for example, some models return “CoNLL-2003 English” for both the train and test datasets, while others report ``CoNLL-2003 English'' for the train dataset and ``CoNLL-2003 English (test)'' for the test dataset. As expected, the performance for the open domain are much lower, as this is a more challenging and realistic setting.

\paragraph{Leaderboard Results} Table~\ref{tab:leaderboard_results} presents the results for leaderboard construction. The best-performing model, Gemini-2.5-Pro, achieves a leaderboard recall of only 49.46 in the closed domain, meaning fewer than half of the gold-standard leaderboards are reconstructed. Paper coverage (41.05), result coverage (36.27), and average overlap (7.99) show that even when some leaderboards are reconstructed, the included papers and results only partially match the gold data, and the ranking similarity is very limited. For \citet{Kardas2020AXCELL}, despite some overlap between the gold and predicted leaderboard, it was very limited (0.00). In the more realistic open-domain setting, leaderboard results drop significantly across all models. Both domains show substantial room for improvement.

\paragraph{Experiment Type Results}
\autoref{tab:results_experiment-type} reports the correctly classifying on the tuples based on \textit{experiment type}. ~\citet{sahinuc2024efficient} achieve the best performance, correctly classifying $\sim$82\% and $\sim$81\% of tuples in the closed and open domains, respectively. One possible explanation for the strong ET-Accuracy of the \citet{sahinuc2024efficient} baseline is that the method may have extracted only a subset of simpler tuples, where the experimental context was easier to infer. Surprisingly, GPT-4.1 and Gemini-1.5-Pro show a big difference between the closed- and open-domain settings.

\paragraph{Error Analysis Experiment Type}

To analyse misclassifications in \textit{experiment type}, we create a confusion matrix. The matrix for closed-domain GPT-4.1 is shown in \autoref{tab:confusion-matrix}; others appear in \autoref{app:experiment-type}. The matrix shows that proposed methods are frequently misclassified as method variations (23.57\%), and vice versa (11.55\%), while baselines are more reliably identified (89.67\% correct). This is likely because authors often use vague phrasing such as “our approach” or “this setting”, and do not clearly distinguish between introducing a single method and presenting multiple variations.

\section{Discussion}\label{sec:discussion}

\paragraph{Additional Metadata} For future work, we aim to expand the scope of the leaderboard by including additional metadata, such as method names and related attributes. As demonstrated by ~\citet{otto2023gsap}, this is a challenging task, as their paper shows that labelling ML models the annotator agreement is low (e.g. for \textit{Model Architecture} the annotator agreement was only 23.7 for exact-match and 34.4 for partial-match). Second, terminology and conceptual frameworks may evolve over time, making consistent annotation difficult across papers from different publication periods.

\paragraph{Improved Extraction Framework} 
While individual entity extraction is strong, combining entities correctly remains challenging, as reflected by leaderboard-specific metrics like leaderboard recall. {\color{green} While this paper only investigates zero-shot extraction, future work could explore whether supervised adaptation improves structured result extraction of tuples such as task, dataset, metric, score, and experiment type from the \textsc{MetaLead} dataset.} In addition, future work could explore prompt engineering to better align the model with annotation guidelines~\cite{sainzgollie, huang2025guidener}. Vision-language models (VLMs), such as \textit{Nanonets}~\citep{Nanonets-OCR-S}, may enhance extraction from tables and figures that are difficult for text-only models. Multi-agent systems also show promise~\cite{song2025scientific}, by dividing extraction into specialised subtasks (e.g., tuple extraction, validation, experiment-type labelling) and cross-verifying outputs for greater consistency and completeness.

\paragraph{Flexible Leaderboards}
Traditional leaderboard schemas have been static, relying on fixed structures. Ideally, users should define their own schema and filter results based on individual needs, for example, viewing only \textit{ID} results, including \textit{OOD} results, or selecting the best result and baseline per paper. This flexibility would make leaderboards more user-centric. Exploring how such dynamic, customisable leaderboards could work in practice presents an interesting direction for future research. 

\paragraph{Benchmarking Negative Results and Model Failures}
\textsc{MetaLead} includes negative results and failed variants, traditionally excluded from traditional leaderboards. This enables systematic analysis of model weaknesses and failure patterns. Future work could develop meta-analysis tools to identify common conditions for poor performance, such as specific task types, dataset setups, or metrics. Highlighting and categorising failures promotes transparent reporting and supports the development of more robust models and evaluation frameworks~\cite{karl2024position}.

\paragraph{Downstream Tasks} \textsc{MetaLead} can be used for a wide range of downstream tasks. The simplest is for a researcher to quickly survey a field, identify gaps, and plan new experiments. More advanced applications include result-centric summarisation, citation recommendation, and trend analysis of datasets and metrics. \textsc{MetaLead} also supports predictive modelling, benchmark coverage auditing, and the construction of structured knowledge graphs (e.g. \citet{kabongo2024orkg}). {\color{green}An important downstream task is model selection for large-scale industry deployment. Traditional leaderboards would only display the best result per paper (e.g. 91.62 and 90.32 F1 for NER in \autoref{fig:before-after-metalead}), but an industry practitioner may also value robustness and cost. Using leaderboard that conforms to the \textsc{MetaLead} data schema, one may notice that a baseline method performs slightly below the top results(e.g., 90.20 F1 for NER). Such information is useful to the industry practitioner, as the baseline method may be cheaper, making it more cost-effective in a production setting.} By including all experimental results with rich metadata, \textsc{MetaLead} can facilitate more transparent, scalable, and automated research workflows.

\section{Conclusion}\label{sec:conclusion}


In this paper, we present \textsc{MetaLead}, a fully human-annotated comprehensive dataset designed to improve the transparency and utility of ML leaderboard reporting. \textsc{MetaLead} addresses key limitations of existing datasets by (1) including \textbf{all reported results}, such as baselines and ablations, rather than only the best ones; (2) categorising each result by \textbf{experiment type} to support structured comparisons; and (3) \textbf{distinguishing train and test datasets} to enable cross-domain assessment. These schema innovations promote more nuanced evaluations, reduce ambiguity, and facilitate downstream tasks such as benchmarking model robustness, auditing evaluation practices, and enabling user-defined leaderboard construction. As demonstrated by the benchmarking results, extracting complete and structured information remains challenging, but \textsc{MetaLead} provides a clear reference standard and supports the development of more rigorous and transparent ML research.
\section*{Limitations}\label{sec:limitations}
Our work presents several important limitations. First, 
{\color{green}although \textsc{MetaLead} contains 3,568 annotated results, an order of magnitude more than SciLead, the number of source papers remains modest. Future extensions should expand beyond 43 ML papers to include other scientific domains and non-English publications, in order to enhance generalisability and coverage.} 
Second, the papers are restricted to the machine learning domain, and the timeframe considered further narrows the applicability of the results. Third, we focus exclusively on papers written in English, which limits the applicability of our methods to non-English scientific literature. Fourth, while a leaderboard could theoretically be expanded to finer levels of granularity, such as more detailed score breakdowns or nuanced experimental conditions, we chose not to pursue this direction. This decision was informed by prior research that demonstrates low inter-annotator agreement when annotating machine learning scores and other experimental details within papers~\cite{otto2023gsap}. Maintaining consistency and ensuring meaningful comparisons across such fine-grained annotations introduces significant challenges. As such, we recommend that researchers refer to the original papers for complete experimental details. Finally, our extraction methods target experimental results presented in the main body of papers. We excluded results presented in figures (e.g. line graphs), as the corresponding values are often ambiguous or too difficult to extract reliably.
\section*{Ethical Considerations}

For our annotations, we followed the instructions from the ethics board of the Commonwealth Scientific and Industrial Research Organisation (CSIRO). The annotations were conducted by the authors as part of their paid research duties. The authors, as domain experts familiar with the \textsc{MetaLead} papers, served as annotators, a common practice in leaderboard research. The dataset does not contain any sensitive information, and no personal information about the annotators was recorded. The \textsc{MetaLead} dataset will be made publicly available for research purposes. Given its focus on scientific literature and the nature of its content, we do not anticipate any risk of misuse for malicious purposes.

\textsc{MetaLead} is built upon the dataset introduced by \citet{sahinuc2024efficient}, which was published under the \textit{CC-BY-NC 4.0 (Attribution-NonCommercial 4.0 International)} licence. In accordance with the licence terms, we attribute the original authors and ensure that our use remains strictly non-commercial. The original dataset served as a foundation, which we substantially expanded through additional manual annotations and revised schema definitions. The combined dataset, including our extensions, will be released solely for academic research purposes. No modifications were made to the original data beyond reformatting for consistency, and the provenance of the original annotations is preserved.

\bibliographystyle{acl_natbib}
\bibliography{anthology,custom}

\clearpage
\newpage
\appendix 

\section{Annotation Guidelines}\label{app:annotation_guidelines}
\begin{figure}[H]
\normalsize
\rule{\linewidth}{0.4pt}
\textbf{Objective:} Identify and extract all experimental results from the paper, including baseline and ablation studies for each specified task. Extract results in the format \textit{$<$Task, Train Dataset, Test Dataset, Metric, Result$>$}, e.g.:
\textit{$<$``Task": ``Language Modeling", ``Train Dataset": ``WikiText-103", ``Test Dataset": ``WikiText-103", ``Metric": ``Perplexity", ``Result": 28.0$>$} or 
\textit{$<$``Task": ``Language Modeling", ``Train Dataset": ``Penn Treebank (PTB)", ``Test Dataset": ``Penn Treebank (PTB)", ``Metric": ``Perplexity", ``Result": 46.01$>$} \\

\textbf{Guidelines:}
\begin{itemize}
    \item {\textbf{Limit to the tasks provided:}} Only report results related to the tasks in the original annotation (no other tasks). However, include results for all datasets and metrics associated with the task, even if the original annotations do not.
    \item \textbf{Avoid duplicates:} If the same result is reported multiple times (e.g., in multiple tables), ensure that you only report it once. 
    \item \textbf{Limit to ML experiments:} Exclude experimental results from human annotators.  
    \item \textbf{Differentiate between train and test datasets:} The training dataset and the test dataset are the same; however, in rare cases, they might differ.
\end{itemize}

\rule{\linewidth}{0.4pt}
\caption{Instructions Part 1 of the Annotation Process.}\label{fig:instructions_annotations1}
\end{figure}

\begin{figure}[!t]
\normalsize
\rule{\linewidth}{0.4pt}

\textbf{Objective:} Label each experimental result based on its type of experiment. \\

\textbf{Guidelines:} Choose one of the following labels:

\begin{enumerate}
    \item \textbf{Baseline comparison:} Results that are benchmarks or comparisons with prior work {\color{black}(e.g. previous state-of-the-art results)}. 
    
    \item \textbf{Proposed method:} Experiments with the main method proposed by the authors. {\color{black}The method proposed by the authors is generally regarded as their main contribution. Authors often refer to this method as ``Our Model" or ``Our Method". If several fine-tuning variants are tested (typically on a development set), the best-performing version that is used for test set evaluation should be labelled as the proposed method, while the others are considered ablations.}
    
    \item \textbf{Ablation study:} Experiments that test the contribution of specific features, modules, hyperparameter tuning, etc. If there are ablation studies for the baselines, label them as baseline comparisons, NOT ablation studies.
    Any hybrid/ensemble method that includes the proposed method should also be labelled as an ablation study. {\color{black} Experiments using the main proposed method, even when conducted on a subset of the data, should be labelled as \textit{proposed method} and should not be classified as ablation studies, unless they do experiments on the subset of the data to determine best features, hyperparameter, etc. Changes to the pre-training setup (but not the train dataset) should also be treated as ablation studies.}
\end{enumerate}

If the authors clearly specify the type of experiment, follow their classification, otherwise, use your own judgement.

\rule{\linewidth}{0.4pt}
\caption{Instructions Part 2 of the Annotation Process.}\label{fig:instructions_annotations2}
\end{figure}

\FloatBarrier

\section{Prompts}\label{sec:appendix_prompts}
Prompt for the first extraction step of the leaderboard tuples is shown in Figure~\ref{fig:prompt_extraction}, and, the normalisation prompt is shown in Figure~\ref{fig:prompt_normalisation}.

\begin{figure}[H]
\rule{\linewidth}{0.4pt}
        \small
        \setstretch{0.8} 
        \texttt{You are given several parts of a research paper as input. \\
  Extract these tuples for all experimental results. \\
        Each tuple must contain:}
        
        \begin{itemize}[leftmargin=0.5em]
            \item[] -- \texttt{a single task, single train dataset, single test dataset, single metric, single numeric result, experiment-type label.}
        \end{itemize}
        
        \texttt{Ensure tuples are not aggregated and contain a single value for each field.}

        \texttt{Use the following JSON format:}

        \setstretch{0.8} 
        \begin{adjustwidth}{1em}{0pt}
        \texttt{\{}\\
        \quad\texttt{``Task": ``Task name",}\\
        \quad\texttt{``Train-Dataset": ``Train dataset name",}\\
        \quad\texttt{``Test-Dataset": ``Test dataset name",}\\
        \quad\texttt{``Metric": ``Metric name",}\\
        \quad\texttt{``Result": ``Result score",}\\
        \quad\texttt{``Experiment-Type": ``Experiment type label (``proposed method'', ``baseline'', or ``variation of proposed method'')"}\\
        \texttt{\}}\\
        \end{adjustwidth}

\vspace{-1em}
\rule{\linewidth}{0.4pt}
    \caption{Prompt for the first extraction step.}
    \label{fig:prompt_extraction}
\end{figure}

\begin{figure}[H]
\rule{\linewidth}{0.4pt}
        \small
        \setstretch{0.8} 

        \texttt{Match the input \{input\_item\} with the best candidate from \{candidates\_list\}. \\
            Return the exact matching item or the input if no match is found.\\
            Only return the exact matching item, nothing else.}\\
\vspace{-1em}
\rule{\linewidth}{0.4pt}
    \caption{Prompt for the normalisation step.}
    \label{fig:prompt_normalisation}
\end{figure}

\FloatBarrier

\section{Experiment Setup Details}\label{sec:appendix_exp_setup_details}

For the proprietary and open source models, we have only selected models with a minimum of 5,000 output token. We use three OpenAI models with the Azure API: \textit{GPT-4.1}, \textit{GPT-4o}, and, \textit{o4-mini}. The approximate cost for all the experiments with the Azure API is approximately USD 25 Of Google, we use three different versions of the Gemini family \textit{Gemini-2.5-Pro}, \textit{Gemini-2.5-Flash}, and, \textit{Gemini-1.5-Pro}. The approximate cost for the Gemini API is USD 50. For Mistral, we use \textit{Mistral-Large-2402}, for which the cost is approximately USD10. For Llama of Meta, we use the Huggingface version of \textit{Llama 3.3 70B}. The experiments for \textit{Llama 3.3 70B} were conducted locally on four Nvidia H100 GPUs. Before passing content to the large language models, all PDFs are parsed using PyMuPDF\footnote{\url{https://github.com/pymupdf/PyMuPDF}}, which preserves the layout structure. We exclude the reference list and the appendix.

\paragraph{Responsible AI Content Filtering}  Some of the normalisation prompts triggered Azure’s content filter, despite containing no harmful or inappropriate material. For instance, prompts that included the metric \textit{V-M} consistently triggered Azure's content filter. As a result, we were required to slightly modify certain prompts to bypass these restrictions. When a prompt triggered the filter, we retained the originally extracted entity without applying further normalisation.

\paragraph{AxCell}
The AxCell baseline of ~\citet{Kardas2020AXCELL}, relies on input files for which source code is available. Consequently, AxCell can only be executed for 36 out of the 43 papers. Due to the way AxCell is originally designed, it can only extract tuples in the format \textit{$\langle$task, dataset, metric, score$\rangle$}. For comparison with the \textsc{MetaLead} dataset, we align AxCell’s extracted “dataset” field with the “test dataset” field in \textsc{MetaLead}. During the normalisation step, we employ the GPT-4.1 model. We have modified the original AxCell codebase to extract all available results, including baselines, ablations, and results with low scores, rather than filtering them out.

\paragraph{SciLead}
For the SciLead baseline of~\cite{sahinuc2024efficient}, we use the GPT-4.1 model for two reasons. First, our own simple prompting experiments with GPT-4.1 yield together with Gemini-2.5-Pro the best performance. Second, in the study of~\citet{sahinuc2024efficient}, the implementation using a GPT-family model achieved the strongest performance.

\FloatBarrier

\section{Experiment-Type Classification Confusion Matrices} \label{app:experiment-type}

\begin{table}[H]
\setlength{\tabcolsep}{2pt}
\centering
\setlength{\aboverulesep}{0.4ex}
\setlength{\belowrulesep}{0.2ex}
\footnotesize
\begin{tabular}{lrrr}
\toprule
\textbf{Gold / Predicted} & \textbf{Baseline} & \makecell[l]{\textbf{Proposed}\\\textbf{Method}} & \makecell[l]{\textbf{Method}\\\textbf{Variation}} \\
\midrule
\textbf{Baseline} & 81.62 & 10.00 & 8.38 \\
\textbf{Proposed Method} & 0.39 & 88.14 & 11.47 \\
\textbf{Method Variation} & 0.00 & 35.71 & 64.29 \\
\bottomrule
\end{tabular}
\caption{Confusion matrix for \textit{experiment type} classification by GPT-4o in the closed domain. Percentages are averaged over 34 papers with non-empty matrices.}
\label{tab:confusion-matrix-gpt4o-closed}
\end{table}

\begin{table}[H]
\setlength{\tabcolsep}{2pt}
\centering
\setlength{\aboverulesep}{0.4ex}
\setlength{\belowrulesep}{0.2ex}
\footnotesize
\begin{tabular}{lrrr}
\toprule
\textbf{Gold / Predicted} & \textbf{Baseline} & \makecell[l]{\textbf{Proposed}\\\textbf{Method}} & \makecell[l]{\textbf{Method}\\\textbf{Variation}} \\
\midrule
\textbf{Baseline} & 97.47 & 0.70 & 1.83 \\
\textbf{Proposed Method} & 0.89 & 70.18 & 28.93 \\
\textbf{Method Variation} & 12.89 & 6.52 & 80.59 \\
\bottomrule
\end{tabular}
\caption{Confusion matrix for \textit{experiment type} classification by o4-mini in the closed domain. Percentages are averaged over 36 papers with non-empty matrices.}
\label{tab:confusion-matrix-o4mini-closed}
\end{table}

\begin{table}[H]
\setlength{\tabcolsep}{2pt}
\centering
\setlength{\aboverulesep}{0.4ex}
\setlength{\belowrulesep}{0.2ex}
\footnotesize
\begin{tabular}{lrrr}
\toprule
\textbf{Gold / Predicted} & \textbf{Baseline} & \makecell[l]{\textbf{Proposed}\\\textbf{Method}} & \makecell[l]{\textbf{Method}\\\textbf{Variation}} \\
\midrule
\textbf{Baseline} & 94.70 & 0.47 & 4.83 \\
\textbf{Proposed Method} & 0.32 & 77.53 & 22.15 \\
\textbf{Method Variation} & 7.41 & 10.20 & 82.39 \\
\bottomrule
\end{tabular}
\caption{Confusion matrix for \textit{experiment type} classification by Gemini-2.5-Pro in the closed domain. Percentages are averaged over 37 papers with non-empty matrices.}
\label{tab:confusion-matrix-gemini25pro-closed}
\end{table}

\begin{table}[H]
\setlength{\tabcolsep}{2pt}
\centering
\setlength{\aboverulesep}{0.4ex}
\setlength{\belowrulesep}{0.2ex}
\footnotesize
\begin{tabular}{lrrr}
\toprule
\textbf{Gold / Predicted} & \textbf{Baseline} & \makecell[l]{\textbf{Proposed}\\\textbf{Method}} & \makecell[l]{\textbf{Method}\\\textbf{Variation}} \\
\midrule
\textbf{Baseline} & 95.83 & 0.76 & 3.41 \\
\textbf{Proposed Method} & 2.50 & 80.00 & 17.50 \\
\textbf{Method Variation} & 14.29 & 0.00 & 85.71 \\
\bottomrule
\end{tabular}
\caption{Confusion matrix for \textit{experiment type} classification by Gemini-2.5-Flash in the closed domain. Percentages are averaged over 23 papers with non-empty matrices.}
\label{tab:confusion-matrix-gemini25flash-closed}
\end{table}

\begin{table}[H]
\setlength{\tabcolsep}{2pt}
\centering
\setlength{\aboverulesep}{0.4ex}
\setlength{\belowrulesep}{0.2ex}
\footnotesize
\begin{tabular}{lrrr}
\toprule
\textbf{Gold / Predicted} & \textbf{Baseline} & \makecell[l]{\textbf{Proposed}\\\textbf{Method}} & \makecell[l]{\textbf{Method}\\\textbf{Variation}} \\
\midrule
\textbf{Baseline} & 87.33 & 11.00 & 1.67 \\
\textbf{Proposed Method} & 5.28 & 71.43 & 23.29 \\
\textbf{Method Variation} & 28.57 & 13.10 & 58.33 \\
\bottomrule
\end{tabular}
\caption{Confusion matrix for \textit{experiment type} classification by Gemini-1.5-Pro in the closed domain. Percentages are averaged over 35 papers with non-empty matrices.}
\label{tab:confusion-matrix-gemini15pro-closed}
\end{table}

\begin{table}[H]
\setlength{\tabcolsep}{2pt}
\centering
\setlength{\aboverulesep}{0.4ex}
\setlength{\belowrulesep}{0.2ex}
\footnotesize
\begin{tabular}{lrrr}
\toprule
\textbf{Gold / Predicted} & \textbf{Baseline} & \makecell[l]{\textbf{Proposed}\\\textbf{Method}} & \makecell[l]{\textbf{Method}\\\textbf{Variation}} \\
\midrule
\textbf{Baseline} & 81.04 & 4.72 & 14.24 \\
\textbf{Proposed Method} & 1.72 & 87.35 & 10.92 \\
\textbf{Method Variation} & 11.33 & 12.02 & 76.64 \\
\bottomrule
\end{tabular}
\caption{Confusion matrix for \textit{experiment type} classification by LLaMA 3.3 70B in the closed domain. Percentages are averaged over 32 papers with non-empty matrices.}
\label{tab:confusion-matrix-llama33-closed}
\end{table}

\begin{table}[H]
\setlength{\tabcolsep}{2pt}
\centering
\setlength{\aboverulesep}{0.4ex}
\setlength{\belowrulesep}{0.2ex}
\footnotesize
\begin{tabular}{lrrr}
\toprule
\textbf{Gold / Predicted} & \textbf{Baseline} & \makecell[l]{\textbf{Proposed}\\\textbf{Method}} & \makecell[l]{\textbf{Method}\\\textbf{Variation}} \\
\midrule
\textbf{Baseline} & 85.89 & 5.59 & 8.52 \\
\textbf{Proposed Method} & 0.00 & 77.51 & 22.49 \\
\textbf{Method Variation} & 15.28 & 16.67 & 68.06 \\
\bottomrule
\end{tabular}
\caption{Confusion matrix for \textit{experiment type} classification by Mistral Large in the closed domain. Percentages are averaged over 37 papers with non-empty matrices.}
\label{tab:confusion-matrix-mistral-closed}
\end{table}

\begin{table}[H]
\setlength{\tabcolsep}{2pt}
\centering
\setlength{\aboverulesep}{0.4ex}
\setlength{\belowrulesep}{0.2ex}
\footnotesize
\begin{tabular}{lrrr}
\toprule
\textbf{Gold / Predicted} & \textbf{Baseline} & \makecell[l]{\textbf{Proposed}\\\textbf{Method}} & \makecell[l]{\textbf{Method}\\\textbf{Variation}} \\
\midrule
\textbf{Baseline} & 67.79 & 2.30 & 11.73 \\
\textbf{Proposed Method} & 4.22 & 58.50 & 14.55 \\
\textbf{Method Variation} & 4.70 & 5.56 & 53.38 \\
\bottomrule
\end{tabular}
\caption{Confusion matrix for \textit{experiment type} classification by \citet{sahinuc2024efficient} in the closed domain. Percentages are averaged over 22 papers with non-empty matrices.}
\label{tab:confusion-matrix-sahinuc-closed}
\end{table}

\begin{table}[H]
\setlength{\tabcolsep}{2pt}
\centering
\setlength{\aboverulesep}{0.4ex}
\setlength{\belowrulesep}{0.2ex}
\footnotesize
\begin{tabular}{lrrr}
\toprule
\textbf{Gold / Predicted} & \textbf{Baseline} & \makecell[l]{\textbf{Proposed}\\\textbf{Method}} & \makecell[l]{\textbf{Method}\\\textbf{Variation}} \\
\midrule
\textbf{Baseline} & 90.05 & 1.07 & 8.88 \\
\textbf{Proposed Method} & 1.33 & 71.07 & 27.59 \\
\textbf{Method Variation} & 7.29 & 6.85 & 85.86 \\
\bottomrule
\end{tabular}
\caption{Confusion matrix for \textit{experiment type} classification by GPT-4.1 in the open domain. Percentages are averaged over 18 papers with non-empty matrices.}
\label{tab:confusion-matrix-open}
\end{table}

\begin{table}[H]
\setlength{\tabcolsep}{2pt}
\centering
\setlength{\aboverulesep}{0.4ex}
\setlength{\belowrulesep}{0.2ex}
\footnotesize
\begin{tabular}{lrrr}
\toprule
\textbf{Gold / Predicted} & \textbf{Baseline} & \makecell[l]{\textbf{Proposed}\\\textbf{Method}} & \makecell[l]{\textbf{Method}\\\textbf{Variation}} \\
\midrule
\textbf{Baseline} & 90.00 & 4.17 & 5.83 \\
\textbf{Proposed Method} & 0.00 & 91.67 & 8.33 \\
\textbf{Method Variation} & 16.67 & 33.33 & 50.00 \\
\bottomrule
\end{tabular}
\caption{Confusion matrix for \textit{experiment type} classification by GPT-4o in the open domain. Percentages are averaged over 34 papers with non-empty matrices.}
\label{tab:confusion-matrix-gpt4o-open}
\end{table}

\begin{table}[H]
\setlength{\tabcolsep}{2pt}
\centering
\setlength{\aboverulesep}{0.4ex}
\setlength{\belowrulesep}{0.2ex}
\footnotesize
\begin{tabular}{lrrr}
\toprule
\textbf{Gold / Predicted} & \textbf{Baseline} & \makecell[l]{\textbf{Proposed}\\\textbf{Method}} & \makecell[l]{\textbf{Method}\\\textbf{Variation}} \\
\midrule
\textbf{Baseline} & 96.75 & 0.48 & 2.77 \\
\textbf{Proposed Method} & 0.59 & 66.00 & 33.41 \\
\textbf{Method Variation} & 12.50 & 9.38 & 78.13 \\
\bottomrule
\end{tabular}
\caption{Confusion matrix for \textit{experiment type} classification by o4-mini in the open domain. Percentages are averaged over 35 papers with non-empty matrices.}
\label{tab:confusion-matrix-o4mini-open}
\end{table}

\begin{table}[H]
\setlength{\tabcolsep}{2pt}
\centering
\setlength{\aboverulesep}{0.4ex}
\setlength{\belowrulesep}{0.2ex}
\footnotesize
\begin{tabular}{lrrr}
\toprule
\textbf{Gold / Predicted} & \textbf{Baseline} & \makecell[l]{\textbf{Proposed}\\\textbf{Method}} & \makecell[l]{\textbf{Method}\\\textbf{Variation}} \\
\midrule
\textbf{Baseline} & 96.48 & 0.49 & 3.04 \\
\textbf{Proposed Method} & 1.14 & 90.44 & 8.42 \\
\textbf{Method Variation} & 7.14 & 2.34 & 90.52 \\
\bottomrule
\end{tabular}
\caption{Confusion matrix for \textit{experiment type} classification by Gemini-2.5-Pro in the open domain. Percentages are averaged over 25 papers with non-empty matrices.}
\label{tab:confusion-matrix-gemini25pro-open}
\end{table}

\begin{table}[H]
\setlength{\tabcolsep}{2pt}
\centering
\setlength{\aboverulesep}{0.4ex}
\setlength{\belowrulesep}{0.2ex}
\footnotesize
\begin{tabular}{lrrr}
\toprule
\textbf{Gold / Predicted} & \textbf{Baseline} & \makecell[l]{\textbf{Proposed}\\\textbf{Method}} & \makecell[l]{\textbf{Method}\\\textbf{Variation}} \\
\midrule
\textbf{Baseline} & 97.94 & 0.00 & 2.06 \\
\textbf{Proposed Method} & 0.00 & 75.10 & 24.90 \\
\textbf{Method Variation} & 9.98 & 10.83 & 79.19 \\
\bottomrule
\end{tabular}
\caption{Confusion matrix for \textit{experiment type} classification by Gemini-2.5-Flash in the open domain. Percentages are averaged over 25 papers with non-empty matrices.}
\label{tab:confusion-matrix-gemini25flash-open}
\end{table}

\begin{table}[H]
\setlength{\tabcolsep}{2pt}
\centering
\setlength{\aboverulesep}{0.4ex}
\setlength{\belowrulesep}{0.2ex}
\footnotesize
\begin{tabular}{lrrr}
\toprule
\textbf{Gold / Predicted} & \textbf{Baseline} & \makecell[l]{\textbf{Proposed}\\\textbf{Method}} & \makecell[l]{\textbf{Method}\\\textbf{Variation}} \\
\midrule
\textbf{Baseline} & 95.83 & 1.14 & 3.03 \\
\textbf{Proposed Method} & 6.25 & 83.33 & 10.42 \\
\textbf{Method Variation} & 33.33 & 5.56 & 61.11 \\
\bottomrule
\end{tabular}
\caption{Confusion matrix for \textit{experiment type} classification by Gemini-1.5-Pro in the open domain. Percentages are averaged over 17 papers with non-empty matrices.}
\label{tab:confusion-matrix-gemini15pro-open}
\end{table}

\begin{table}[H]
\setlength{\tabcolsep}{2pt}
\centering
\setlength{\aboverulesep}{0.4ex}
\setlength{\belowrulesep}{0.2ex}
\footnotesize
\begin{tabular}{lrrr}
\toprule
\textbf{Gold / Predicted} & \textbf{Baseline} & \makecell[l]{\textbf{Proposed}\\\textbf{Method}} & \makecell[l]{\textbf{Method}\\\textbf{Variation}} \\
\midrule
\textbf{Baseline} & 85.61 & 5.17 & 9.22 \\
\textbf{Proposed Method} & 1.57 & 86.61 & 11.82 \\
\textbf{Method Variation} & 10.50 & 15.49 & 74.00 \\
\bottomrule
\end{tabular}
\caption{Confusion matrix for \textit{experiment type} classification by LLaMA-3.3-70B-Instruct in the open domain. Percentages are averaged over 35 papers with non-empty matrices.}
\label{tab:confusion-matrix-llama33-open}
\end{table}

\begin{table}[H]
\setlength{\tabcolsep}{2pt}
\centering
\setlength{\aboverulesep}{0.4ex}
\setlength{\belowrulesep}{0.2ex}
\footnotesize
\begin{tabular}{lrrr}
\toprule
\textbf{Gold / Predicted} & \textbf{Baseline} & \makecell[l]{\textbf{Proposed}\\\textbf{Method}} & \makecell[l]{\textbf{Method}\\\textbf{Variation}} \\
\midrule
\textbf{Baseline} & 79.82 & 4.94 & 15.24 \\
\textbf{Proposed Method} & 0.83 & 67.42 & 31.75 \\
\textbf{Method Variation} & 12.94 & 16.43 & 70.63 \\
\bottomrule
\end{tabular}
\caption{Confusion matrix for \textit{experiment type} classification by Mistral Large in the open domain. Percentages are averaged over 38 papers with non-empty matrices.}
\label{tab:confusion-matrix-mistral-open}
\end{table}

\begin{table}[H]
\setlength{\tabcolsep}{2pt}
\centering
\setlength{\aboverulesep}{0.4ex}
\setlength{\belowrulesep}{0.2ex}
\footnotesize
\begin{tabular}{lrrr}
\toprule
\textbf{Gold / Predicted} & \textbf{Baseline} & \makecell[l]{\textbf{Proposed}\\\textbf{Method}} & \makecell[l]{\textbf{Method}\\\textbf{Variation}} \\
\midrule
\textbf{Baseline} & 88.89 & 2.78 & 8.33 \\
\textbf{Proposed Method} & 12.50 & 75.00 & 12.50 \\
\textbf{Method Variation} & 6.25 & 0.00 & 18.75 \\
\bottomrule
\end{tabular}
\caption{Confusion matrix for \textit{experiment type} classification by \citet{sahinuc2024efficient} in the open domain. Percentages are averaged over 4 papers with non-empty matrices.}
\label{tab:confusion-matrix-sahinuc-open}
\end{table}

\FloatBarrier
\section{Unique Entities in the \textsc{MetaLead}}\label{appendix:unique-entities}

\begin{figure}[H]
\rule{\linewidth}{0.4pt}
\textit{'AVG', 'Accuracy', 'BERTScore', 'BLEU', 'BLEU-4', 'Exact Match (EM)', 'F-Score (F-S)', 'Fuzzy B-Cubed (FBC)', 'Fuzzy normalized mutual information (FNMI)', 'Labeled Attachment Score', 'METEOR', 'Matthew's Correlation Coefficient (MCC)', 'NIST-4', 'Overall-Accuracy', 'Perplexity', 'Precision', 'ROUGE-1', 'ROUGE-2', 'ROUGE-L', 'Recall', 'Sent-Accuracy', 'Spearman Correlation', 'TER', 'Unlabeled Attachment Score', 'V-Measure (V-M)'}
\rule{\linewidth}{0.4pt}
\caption{The unique metrics in \textsc{MetaLead}}
\end{figure}

\begin{figure}[H]
\rule{\linewidth}{0.4pt}
The 23 unique tasks in \textsc{MetaLead} are: \textit{
'Dialogue Generation', 'Text Chunking', 'Combinatory Categorial Grammar (CCG) Supertagging', 'Part-of-Speech (POS) Tagging', 'Question Answering', 'Relation Classification', 'Text Similarity', 'Dependency Parsing', 'Named Entity Recognition (NER)', 'Constituency Parsing', 'Intent Detection and Slot Filling', 'Question Generation', 'Paraphrase Detection', 'Entity Typing', 'Natural Language Inference (NLI)', 'Sentiment Analysis', 'Summarization', 'Response Generation', 'Machine Translation', 'Dialogue Act Classification', 'Word Sense Induction', 'Linguistic Acceptability', 'Language Modeling'.}
\rule{\linewidth}{0.4pt}
\caption{The unique tasks in \textsc{MetaLead}}
\end{figure}

\begin{figure*}
\rule{\linewidth}{0.4pt}\textit{
'AESLC', 'ATIS', 'AX', 'BC5CDR', 'BIGPATENT', 'BillSum', 'CBS SciTech News', 'CCGBank', 'CCGBank Dev', 'CNN/DailyMail', 'CNN/DailyMail Dev', 'CoLA', 'CoLA Dev', 'CoNLL++', 'CoNLL-03 - English', 'CoNLL-2000', 'CoNLL-2000 Dev', 'CoNLL-2002 - Dutch', 'CoNLL-2002 - Dutch Dev IV', 'CoNLL-2002 - Dutch Dev OOBV', 'CoNLL-2002 - Dutch Dev OOEV', 'CoNLL-2002 - Dutch Dev OOTV', 'CoNLL-2002 - Dutch Test IV', 'CoNLL-2002 - Dutch Test OOBV', 'CoNLL-2002 - Dutch Test OOEV', 'CoNLL-2002 - Dutch Test OOTV', 'CoNLL-2002 - Spanish', 'CoNLL-2002 - Spanish Dev IV', 'CoNLL-2002 - Spanish Dev OOBV', 'CoNLL-2002 - Spanish Dev OOEV', 'CoNLL-2002 - Spanish Dev OOTV', 'CoNLL-2002 - Spanish Test IV', 'CoNLL-2002 - Spanish Test OOBV', 'CoNLL-2002 - Spanish Test OOEV', 'CoNLL-2002 - Spanish Test OOTV', 'CoNLL-2003 - English', 'CoNLL-2003 - English Dev', 'CoNLL-2003 - English Dev IV', 'CoNLL-2003 - English Dev OOBV', 'CoNLL-2003 - English Dev OOEV', 'CoNLL-2003 - English Dev OOTV', 'CoNLL-2003 - English IV', 'CoNLL-2003 - English OOBV', 'CoNLL-2003 - English OOEV', 'CoNLL-2003 - English OOTV', 'CoNLL-2003 - English Test IV', 'CoNLL-2003 - English Test OOBV', 'CoNLL-2003 - English Test OOEV', 'CoNLL-2003 - English Test OOTV', 'CoNLL-2003 - English Train', 'CoNLL-2003 - German', 'CoNLL-2003 - German Dev IV', 'CoNLL-2003 - German Dev OOBV', 'CoNLL-2003 - German Dev OOEV', 'CoNLL-2003 - German Dev OOTV', 'CoNLL-2003 - German Test IV', 'CoNLL-2003 - German Test OOBV', 'CoNLL-2003 - German Test OOEV', 'CoNLL-2003 - German Test OOTV', 'CoQA Dev', 'Connl-2003 - English Dev', 'ConvAI2', 'ConvAI2 Dev', 'DSTC7', 'Dailydialog (DyDA)', 'E-commerce', 'ELI5', 'ELI5 Dev', 'Genia', 'Gigaword', 'Gigaword 10K examples', 'Google Billion Word', 'Google Billion Word Valid', 'ICSI Meeting Recorder Dialog Act Corpus (MRDA)', 'IWSLT’14 EN-DE', 'IWSLT’15 DE-EN', 'IWSLT’15 DE-EN Dev', 'IWSLT’15 EN-VI', 'MNLI-m', 'MNLI-m Dev', 'MNLI-mm', 'MRPC', 'MRPC Dev', 'MapTask', 'Meeting Recorder Dialog Act Corpus (MRDA)', 'Multi-News', 'Multi-domain Food', 'Multi-domain Home', 'Multi-domain Movie', 'Multimodal', 'NCBI', 'NEWSROOM', 'New York Times (NYT)', 'Newsroom', 'OConnl-2003 - English', 'OntoNotes v5 - English', 'OntoNotes v5 - English Dev', 'Ontonotes v4 - Chinese', 'Ontonotes v5 - Englis Dev', 'Ontonotes v5 - English', 'Ontonotes v5 - English BC', 'Ontonotes v5 - English BN', 'Ontonotes v5 - English Dev', 'Ontonotes v5 - English MZ', 'Ontonotes v5 - English NW', 'Ontonotes v5 - English TC', 'Ontonotes v5 - English WB', 'Open Entity', 'Penn Treebank (PTB)', 'Penn Treebank (PTB) Dev', 'Penn Treebank (PTB) Dev IV', 'Penn Treebank (PTB) Dev OOBV', 'Penn Treebank (PTB) Dev OOEV', 'Penn Treebank (PTB) Dev OOTV', 'Penn Treebank (PTB) IV', 'Penn Treebank (PTB) OOBV', 'Penn Treebank (PTB) OOEV', 'Penn Treebank (PTB) OOTV', 'Penn Treebank (PTB) Train', 'PubMed', 'QNLI', 'QNLI Dev', 'QQP', 'QQP Dev', 'RTE', 'RTE Dev', 'ReCoRD', 'ReCoRD dev', 'Reddit TIFU', 'Resume - Chinese', 'Ritter Twitter', 'SNIPS', 'SQuAD 1.1', 'SQuAD 1.1 Dev', 'SQuAD 1.1 dev', 'SQuAD 2.0', 'SQuAD 2.0 Dev', 'SST-2', 'SST-2 Dev', 'STS-B', 'STS-B Dev', 'SemEval 2010 Task 14', 'SemEval 2013 Task 13', 'SemEval 2013 Task 13 with data enrichment', 'Switchboard Dialog Act Corpus (SWDA)', 'TA-CRED', 'WMT’13 EN-DE', 'WMT’14 EN-DE', 'WMT’14 EN-FR', 'WMT’16 RO-EN', 'WMT’17 EN-ZH', 'WNLI', 'WNUT-16 - English', 'WNUT-17 - English', 'Weibo - Chinese', 'WikiHow', 'WikiText-103', 'WikiText-103 Dev', 'Wikitext-2', 'Wikitext-2 Dev', 'XSum', 'XSum Dev', 'arXiv'}
\rule{\linewidth}{0.4pt}
\caption{The unique train and test datasets in \textsc{MetaLead}}
\end{figure*}

\clearpage
\newpage

\end{document}